\documentclass[11pt]{article}
\usepackage{acl2016}
\usepackage{times}
\usepackage{latexsym}
\usepackage{color}
\usepackage{colortbl}
\usepackage [english]{babel}
\usepackage [autostyle, english = american]{csquotes}
\MakeOuterQuote{"}
\usepackage[colorlinks]{}
\usepackage{pgfplots,wrapfig}
\usepackage{tikz}
\usepackage{color}
\usepackage{booktabs}
\usepackage{url}

\newcommand{\ra}[1]{\renewcommand{\arraystretch}{#1}}
\newcolumntype{P}[1]{>{\centering\arraybackslash}p{#1}}
\pgfplotsset{compat=1.12}
\aclfinalcopy

\title{Creating and Characterizing a Diverse Corpus of  Sarcasm in Dialogue} 

   \author{
   \textbf{Shereen Oraby$^*$, Vrindavan Harrison$^*$, Lena Reed$^*$, Ernesto Hernandez$^*$,} \\
   \textbf{Ellen Riloff $^\dag$ and Marilyn Walker$^*$} \\
   $^*$ University of California, Santa Cruz \\ {\tt \{soraby,vharriso,lireed,eherna23,mawalker\}@ucsc.edu} \\
   $^\dag$ University of Utah \\ {\tt riloff@cs.utah.edu}  \\ 
   }

\date{}

\begin{document}

\maketitle

\begin{abstract}
The use of irony and sarcasm in social media allows us to study them
at scale for the first time. However, their diversity
has made it difficult to construct a high-quality corpus of sarcasm in
dialogue. Here, we describe the process of creating 
a large-scale, highly-diverse corpus of online debate forums dialogue, and our novel  methods for
operationalizing classes of sarcasm in the form of rhetorical questions and
hyperbole. We  show that we can use lexico-syntactic cues to reliably retrieve sarcastic utterances with high accuracy.  To demonstrate the properties and quality
of our corpus, we conduct supervised learning experiments with simple
features, and show that we achieve both higher precision and F 
than previous work on sarcasm in debate forums dialogue. We apply a
weakly-supervised linguistic pattern learner and  qualitatively analyze
the linguistic differences in each class.\\

\end{abstract}


\section{Introduction}

Irony and sarcasm in dialogue constitute a highly creative use of
language signaled by a large range of situational,
semantic, pragmatic and lexical cues. Previous work draws
attention to the use of both hyperbole and rhetorical questions in
conversation as distinct types of lexico-syntactic cues defining diverse classes of sarcasm \cite{Gibbs00}.

Theoretical models posit that a single semantic basis underlies
sarcasm's diversity of form, namely "a contrast" between expected and
experienced events, giving rise to a contrast between what is said and
a literal description of the actual situation
\cite{ColstonObrien00b,Partington07}. This semantic characterization
has not been straightforward to operationalize computationally for
sarcasm in dialogue. \newcite{Riloffetal13} operationalize this notion
for sarcasm in tweets, achieving good results. \newcite{Joshietal15}
develop several {\it incongruity} features to capture it, but although 
they improve performance on tweets, their
features do not yield improvements for dialogue.

Previous work on the Internet Argument Corpus (IAC) 1.0 dataset aimed to develop a
high-precision classifier for sarcasm in order to bootstrap a much
larger corpus \cite{LukinWalker13}, but was only able to obtain a
precision of just 0.62, with a best F of 0.57, not high enough for
bootstrapping
\cite{RiloffWiebe03,ThelenRiloff02}. \newcite{Justoetal14}
experimented with the same corpus, using supervised learning, and
achieved a best precision of 0.66 and a best F of
0.70. \newcite{Joshietal15}'s {\it explicit congruity} features
achieve precision around 0.70 and best F of 0.64 on a subset of IAC 1.0.

We decided that we need a larger and more diverse corpus of sarcasm in
dialogue.  It is difficult to efficiently gather sarcastic data, because only about 12\% of the
utterances in written online debate forums dialogue are sarcastic \cite{Walkeretal12c}, and
it is difficult to achieve high
reliability for sarcasm annotation
\cite{Filatova12,Swansonetal14,Gonzalezetal11,Wallaceetal14}. Thus, our contributions are:





\begin{itemize}  \setlength\itemsep{0.1em}
\item We develop a new larger corpus, using several methods that filter
  non-sarcastic utterances to skew the distribution toward/in favor of
  sarcastic utterances. We put filtered data out for annotation, 
  and are able to achieve high annotation reliability. 
\item We present a novel operationalization of both
  rhetorical questions and hyperbole to develop  subcorpora to explore
  the differences between them and general sarcasm.
\item We show that our new corpus is of high quality by applying
supervised machine learning with simple features to explore how different corpus properties 
affect classification results.
We achieve a highest precision of 0.73 and a highest F of
  0.74 on the new corpus with basic n-gram and Word2Vec features, 
 showcasing the quality of the corpus, and improving on previous work.
\item We apply a weakly-supervised learner to characterize linguistic
  patterns in each corpus, and describe the differences across
  generic sarcasm, rhetorical questions and hyperbole in terms of the patterns
  learned.
\item We show for the first time that it is straightforward to develop
  very high precision classifiers for {\sc not-sarcastic} utterances
  across our rhetorical questions and hyperbole subtypes, due to the nature of these utterances
  in debate forum dialogue.
\end{itemize}


\section{Creating a Diverse Sarcasm Corpus}
\label{corp-sec}

There has been relatively little theoretical work on sarcasm in
dialogue that has had access to a large corpus of naturally occurring
examples. \newcite{Gibbs00} analyzes a corpus of 62 conversations
between friends and argues that a robust theory of verbal irony must
account for the large diversity in form. He defines several subtypes,
including rhetorical questions and hyperbole:
\begin{itemize}
  \setlength\itemsep{-0.3em}
\item \textbf{Rhetorical Questions:} asking a question that implies a humorous or critical assertion
\item \textbf{Hyperbole:} expressing a non-literal meaning by exaggerating the reality of a situation
\end{itemize}

Other categories of irony defined by \newcite{Gibbs00} include understatements, jocularity, and 
sarcasm (which he defines as a critical/mocking form of irony). Other work has also tackled jocularity and humor, using different approaches for data aggregation, including filtering by Twitter hashtags, or analyzing laugh-tracks from recordings \cite{Reyesetal12,Bertero16}.

Previous work has not, however, attempted to operationalize these
subtypes in any concrete way. Here we describe our methods for
creating a corpus for generic sarcasm (Gen) (Sec.~\ref{generic-sarc}),
rhetorical questions (RQ), and hyperbole (Hyp)
(Sec.~\ref{rq-hyp-sec}) using data from the Internet Argument Corpus (IAC 2.0).\footnote{The IAC 2.0 is available at https://nlds.soe.ucsc.edu/iac2, and our sarcasm corpus will be released at https://nlds.soe.ucsc.edu/sarcasm2.} Table \ref{table:examp-fig} provides
examples of {\sc sarcastic} and {\sc not-sarcastic} posts from the
corpus we create.  Table~\ref{new-corpus-stats} summarizes the final
composition of our sarcasm corpus.

\begin{table}[t!bh]
\begin{scriptsize}
\centering
\ra{1.3}
\begin{tabular}{@{}p{0.1cm}|p{0.3cm}|p{6.4cm}@{}}\toprule
\multicolumn{3}{c}{\bf \cellcolor[gray]{0.9}Generic Data}  \\
\midrule
1 &$S$ & {I love it when you bash people for stating opinions and no facts when 
you turn around and do the same thing [...] give me a break}  \\ \midrule
2&$NS$ & {The attacker is usually armed in spite of gun control laws. All they do is disarm the law abiding. 
Not to mention the lack of enforcement on criminals.} \\ \bottomrule
\multicolumn{3}{c}{\bf \cellcolor[gray]{0.9}Rhetorical Questions}  \\
\midrule
3&$S$ & \bf {Then why do you call a politician who ran such
measures liberal?} {\it OH yes, it's because you're a 
republican and you're not conservative at all.} \\ \midrule
4&$NS$ & {\bf And what would that prove?} {\it It would certainly show that an animal adapted to 
survival above the Arctic circle was not adapted to the Arizona desert.}\\
\toprule
\multicolumn{3}{c}{\bf \cellcolor[gray]{0.9}Hyperbole}  \\
\midrule

5&$S$ & Thank you for making my point {\bf better than I could ever do!!} It's all 
about you, right honey? I am woman hear me roar right? LMAO \\ \midrule
6&$NS$ & {{\bf Again i am astounded} by the fact that you think i will endanger children. it is a topic sunset, so why are you calling me demented and sick. } \\
\bottomrule
\end{tabular}
\caption{\label{table:examp-fig} {Examples of different types  of {\sc sarcastic} ($S$) and {\sc not-sarcastic} ($NS$) Posts  \\\label{examp-fig}}}
\end{scriptsize}
\end{table}

\begin{table}[t!bh]
\begin{small}
\centering
\ra{1.3}
\begin{tabular}{@{}P{3.4cm}|P{1.35cm}|P{2.1cm}@{}}\toprule
\bf Dataset & \bf  Total Size & \bf Posts Per Class \\ \midrule 
Generic (Gen)  & 6,520 & 3,260 \\ \midrule 
Rhetorical Questions (RQ)  & 1,702 & 851 \\ \midrule 
Hyperbole (Hyp) & 1,164 & 582 \\ 
 \bottomrule
 \end{tabular}
\end{small}
\caption{\label{new-corpus-stats} Total number of posts in each subcorpus (each with a 50\% split of {\sc sarcastic} and {\sc not-sarcastic} posts)}
\vspace{-0.2in}
\end{table}

\subsection{Generic Dataset (Gen)}
\label{generic-sarc}

We first replicated the pattern-extraction experiments of \newcite{LukinWalker13} on
their dataset using AutoSlog-TS \cite{Riloff96}, a weakly-supervised
pattern learner that extracts lexico-syntactic patterns associated with
the input data. We set up the learner to extract patterns for both
{\sc sarcastic} and {\sc not-sarcastic} utterances.  Our first
discovery is that we can classify {\sc not-sarcastic} posts with very
high precision, ranging between 80-90\%.\footnote{We delay a detailed
  discussion of the characteristics of this {\sc not-sarcastic}
  classifier, and the patterns that we learn, until
  Sec.~\ref{ling-anal} where we describe AutoSlog-TS and the
  linguistic characteristics of the whole corpus.}

Because our main goal is to build a larger, more diverse corpus of
sarcasm, we use the high-precision {\sc not-sarcastic}
patterns extracted by AutoSlog-TS to create a "not-sarcastic" filter.
We did this by randomly selecting a new set of 30K posts (restricting to posts with between 10 and 150 words) from IAC 2.0
\cite{Abbottetal16}, and applying the high-precision {\sc
  not-sarcastic} patterns from AutoSlog-TS to filter out any posts that
contain at least one {\sc not-sarcastic} cue. We end up filtering out
two-thirds of the pool, only keeping posts that did not contain any of
our high-precision {\sc not-sarcastic} cues.  We acknowledge that 
this may also filter out sarcastic posts, but we expect
it to increase the ratio of sarcastic posts in the remaining pool. 

\begin{figure}[h]
\begin{center}
   \frame{ \includegraphics[width=3in]{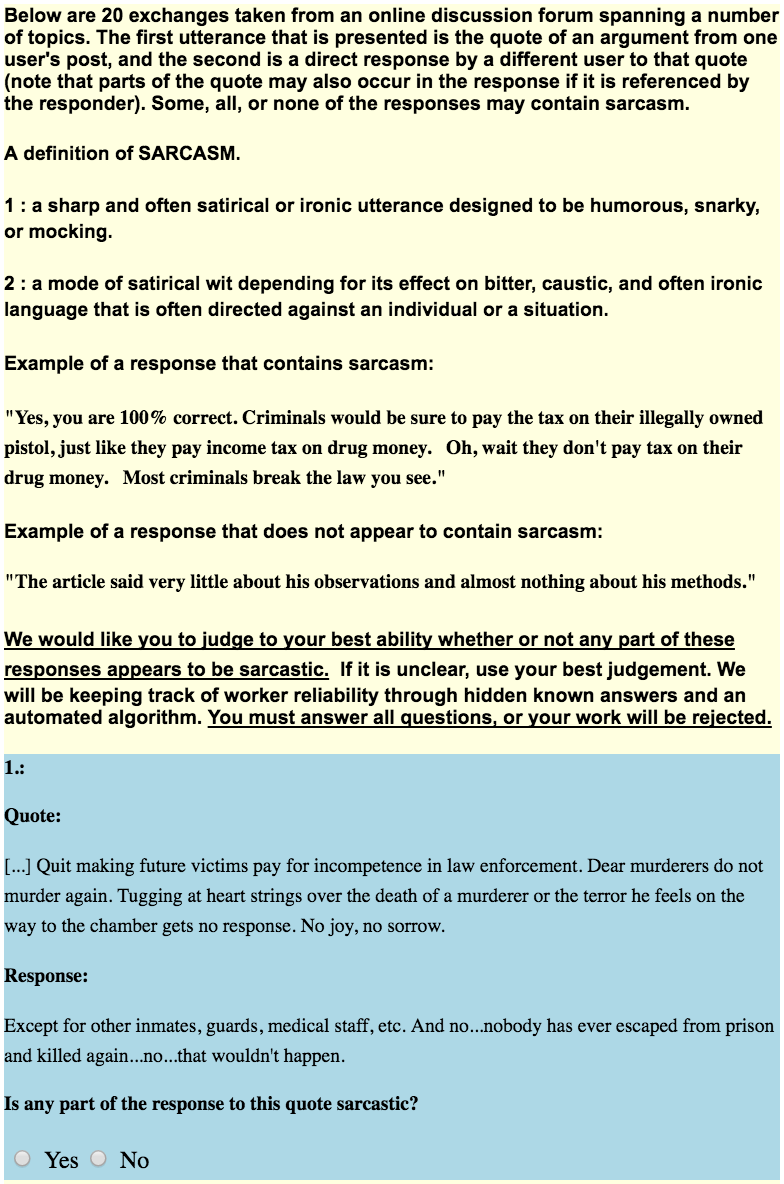}}
  \caption{Mechanical Turk Task Layout}
  \label{hit-fig}
  \end{center}
\end{figure}

We put out the remaining 11,040 posts on Mechanical Turk. As in \newcite{LukinWalker13}, 
we present the posts in "quote-response" pairs, where
the response post to be annotated is presented in the context of its
``dialogic parent'', another post earlier in the thread, or a quote
from another post earlier in the thread \cite{Walkeretal12d}.  In the task instructions, annotators
are presented with a definition of sarcasm, followed by one example of a
quote-response pair that clearly contains sarcasm, and one pair that clearly does not. Each task consists of 20 quote-response pairs that follow the instructions. Figure \ref{hit-fig}
  shows the instructions and layout of a single quote-response pair presented
  to annotators. As in \newcite{LukinWalker13}
and \newcite{Walkeretal12d}, 
annotators are asked a binary question: {\it Is any part of the
  response to this quote sarcastic?}. 

To help filter out unreliable annotators, we
create a qualifier consisting of a set of 20 manually-selected quote-response pairs (10 that should receive a {\sc sarcastic} label and 10 that should receive a {\sc not-sarcastic} label). A Turker must pass the qualifier with a score above 70\% to participate
in our sarcasm annotations tasks.  
    
Our baseline ratio of sarcasm in online debate forums dialogue
is the estimated 12\% sarcastic posts in the IAC, which was found previously by Walker et al.
by gathering annotations for sarcasm, agreement, emotional language, attacks, and nastiness from a subset of around 20K posts from the IAC across various topics \cite{Walkeretal12c}.
Similarly, in his study of recorded conversation among friends, Gibbs cites 8\% sarcastic utterances
among all conversational turns \cite{Gibbs00}.


We choose a conservative threshold: a post is only added to the sarcastic set if at least 6 out of 9 
annotators labeled it sarcastic. Of the 11,040 posts we put out for annotation, we 
thus obtain 2,220 new posts, giving us a ratio of about 20\% sarcasm --
significantly higher than our baseline of 12\%. We choose this conservative threshold to ensure
the quality of our annotations, and we leave aside posts that 5 out of 9 annotators
label as sarcastic for future work -- noting that we can get even higher ratios of sarcasm by including them
(up to 31\%). The percentage agreement between each
annotator and the majority vote is 80\%.

We then expand this set, using only 3 highly-reliable Turkers (based on our
first round of annotations), giving them an
exclusive sarcasm qualification to do additional HITs. We gain 
an additional 1,040 posts for each class when using majority agreement (at least 2 out of 3 sarcasm labels) for the
additional set (to add to the 2,220 original posts).  The average percent agreement with the majority vote
is 89\% for these three annotators. We supplement our sarcastic data with 2,360 not-sarcastic posts from the original data
by \cite{LukinWalker13} that follow our 150-word length restriction, and complete the set with 900 posts that were filtered out by our {\sc not-sarcastic} filter\footnote{We use these unbiased not-sarcastic data sources to avoid using posts coming from the sarcasm-skewed distribution.} -- resulting in a total of 3,260 posts per class (6,520 total posts).

Rows 1 and 2 of Table~\ref{examp-fig} show examples of posts that are labeled sarcastic in our final 
generic sarcasm set. Using our filtering method, we are able to reduce the number of posts
annotated from our original 30K to around 11K, achieving a percentage of
20\% sarcastic posts, even though we choose to use a
conservative threshold of at least 6 out of 9 sarcasm labels. Since the number of
posts being annotated is only a third of the original set size, this method reduces
annotation effort, time, and cost, and helps us shift the distribution of sarcasm
to more efficiently expand our dataset than would otherwise be possible.

\subsection{Rhetorical Questions and Hyperbole}
\label{rq-hyp-sec}

The goal of collecting additional corpora for rhetorical questions and
hyperbole is to increase the diversity of the corpus, and to allow us
to explore the semantic differences between {\sc sarcastic} and {\sc
  not-sarcastic} utterances when particular lexico-syntactic cues are
held constant. We hypothesize that identifying surface-level cues
that are instantiated in {\it both} sarcastic and not sarcastic posts will force
learning models to find deeper semantic cues to distinguish between the classes.
 
Using a combination of findings in the theoretical
literature, and observations of sarcasm patterns in our generic set,
we developed a regex pattern matcher that runs against the 400K
unannotated posts in the IAC 2.0 database and retrieves matching
posts, only pulling posts that have parent posts and a maximum of 150
words.  Table~\ref{table:cue-annotation} only shows a small subset of
the ``more successful'' regex patterns we defined for each class.

\begin{table}[t!bh]
\begin{scriptsize}
\centering
\ra{1.3}
\begin{tabular}{@{}p{3.4cm}|P{0.9cm}|P{0.9cm}|P{1cm}@{}}\toprule
\bf Cue &  \bf \# Found & \bf \# Annot & \bf \% Sarc \\ \hline 
\multicolumn{4}{l}{\bf \cellcolor[gray]{0.9}  Hyperbole}  \\\hline
    \texttt{let's all} & 27 & 21 & 62\% \\ \hline
 \texttt{i love it when} & 158 & 25 & 56\%\\ \hline
   \texttt{oh yeah} & 397 &104 & 50\%\\ \hline
    \texttt{wow} & 977 & 153 & 44\%\\ \hline
     \texttt{i'm * shocked|amazed|impressed} & 120 & 33 & 42\% \\ \hline
       \texttt{fantastic} & 257 & 47 & 36\% \\ \hline
   \texttt{hun/dear*/darling} & 661 & 249 & 32\% \\ \hline
 \texttt{you're kidding/joking} & 132 & 43 & 28\%\\ \hline
 \texttt{eureka} & 21 & 12 & 17\% \\ \hline

\multicolumn{4}{l}{\bf \cellcolor[gray]{0.9}  Rhetorical Questions and Self-Answering}  \\\hline
     \texttt{oh wait} & 136 & 121 & 87\% \\ \hline
    \texttt{oh right} & 19 & 11 & 81\% \\ \hline
     \texttt{oh really} & 62 & 50 & 50\% \\ \hline
    \texttt{really?} & 326 & 151 & 30\% \\ \hline
      \texttt{interesting.} & 48 & 27 & 15\%\\ 
 \bottomrule
 \end{tabular}
\end{scriptsize}
\caption{Annotation Counts for a Subset of Cues \label{table:cue-annotation} \\}
\end{table}

\noindent{\bf Cue annotation experiments.} After running a large
number of retrieval experiments with our regex pattern matcher, we 
select batches of the resulting posts that mix different cue classes
to put out for annotation, in such a way as to not allow the annotators
to determine what regex cues were used. We then
 successively put
out various batches for annotation by 5 of our
highly-qualified annotators, in order to determine what percentage of
posts with these cues are sarcastic.

Table~\ref{table:cue-annotation} summarizes the results for a sample
set of cues, showing the number of posts found containing the cue, the
subset that we put out for annotation, and the percentage of posts
labeled sarcastic in the annotation experiments. For example,
for the hyperbolic cue {\it "wow"}, 977 utterances with the cue were found,
153 were annotated, and 44\% of those were found to be sarcastic (i.e. 56\% were found to be
not-sarcastic). Posts with the cue {\it {"oh wait"}} had the highest sarcasm ratio, at 87\%. 
It is the distinction between the sarcastic and not-sarcastic instances
that we are specifically interested in. We describe the
corpus collection process for each subclass below. 

It is important to note that using particular cues (regex) to retrieve
sarcastic posts {\bf does not} result in posts whose only cue is the
regex pattern. We demonstrate this quantitatively in
Sec.~\ref{ling-anal}. Sarcasm is characterized by
multiple lexical and morphosyntactic cues: these include the use of
intensifiers, elongated words, quotations, false politeness, negative
evaluations, emoticons, and tag questions {\it inter alia}. Table
\ref{table:multiple-cues} shows how sarcastic utterances often contain
combinations of multiple indicators, each playing a role in the overall
sarcastic tone of the post.

\begin{table}[t!bh]
\begin{scriptsize}
\centering
\ra{1.3}
\begin{tabular}{@{}p{7.5cm}@{}}\toprule
 \cellcolor[gray]{0.9} \bf Sarcastic Utterance \\  \hline
Forgive me if I \textbf{doubt} your sincerity, but you seem like \textbf{a troll} 
 to me. \textbf{I suspect} that you aren't interested in learning about  
 evolution {\bf at all}. Your questions, while they do support your  
claim to \textbf{know almost} \textbf{nothing}, are \textbf{pretty} typical of 
creationist \textbf{``prove it to me``} questions.\\ \hline

 \textbf{Wrong again!} You \textbf{obviously} can't recognize refutation when  
 its printed before you. I haven't made the tag \textbf{``you liberals``}  
 derogatory. You liberals have done that to yourselves!  
\textbf{I suppose} you'd rather be called a social reformist! Actually,   
socialist is closer to a true description.  \\
  \bottomrule
\end{tabular}
\caption{Utterances with Multiple Sarcastic Cues \\ \label{table:multiple-cues}}
\end{scriptsize}
\end{table}

\noindent{\bf Rhetorical Questions.}  There is no previous work on
distinguishing sarcastic from non-sarcastic uses of rhetorical
questions (RQs).  RQs are syntactically formulated as a question, but
function as an indirect assertion \cite{Frank90}. The polarity of the
question implies an assertion of the opposite polarity, e.g.  {\it Can
  you read?} implies {\it You can't read.}  RQs are prevalent in
persuasive discourse, and are frequently used ironically
\cite{schaffer-rqs,ilie1994else,Gibbs00}.  Previous work focuses on
their formal semantic properties \cite{Han1a}, or distinguishing RQs
from standard questions \cite{Bhattasalietal15}.


We hypothesized that we could find RQs in abundance by searching for
questions in the middle of a post, that are followed by a
statement, using the assumption that questions followed by a statement
are unlikely to be standard information-seeking questions. We test
this assumption by randomly extracting 100 potential RQs as per our definition and
putting them out on Mechanical Turk to 3 annotators, asking them
whether or not the questions (displayed with their following statement) were
rhetorical. According to majority vote, 75\% of the posts were
rhetorical. 

We thus use this "middle of post" heuristic to
obviate the need to gather manual annotations for RQs, and developed regex patterns to find RQs that 
were more likely to be sarcastic.  A sample of the patterns, number of matches in the corpus, 
the numbers we had annotated, and the percent that are sarcastic after annotation are summarized in
Table~\ref{table:cue-annotation}.

\begin{table}[t!bh]
\begin{scriptsize}
\centering
\ra{1.3}
\begin{tabular}{@{}p{7.5cm}@{}}\toprule
 \cellcolor[gray]{0.9} \bf Rhetorical Questions and Self-Answering\\  \hline
{\bf So you do not wish to have a logical debate?} {\it Alrighty then.
god bless you anyway, brother. }\\ \hline
 {\bf Prove that?} {\it You can't prove that i've given nothing but 
insults. i'm defending myself, to mackindale, that's all.} {\bf do 
 you have a problem with how i am defending myself   
 against mackindale?} {\it Apparently.} \\
\bottomrule
\end{tabular}
\caption{Examples of Rhetorical Questions and Self-Answering \\ \label{table:rqs}}
\end{scriptsize}
\end{table}

We extract 357 posts following the intermediate question-answer pairs
heuristic from our generic (Gen) corpus. We then supplement these with
posts containing RQ cues from our cue-annotation experiments: posts that received 3 out of 5 sarcastic labels in the experiments were considered sarcastic, and posts that received 2 or fewer sarcastic labels were considered not-sarcastic. Our final rhetorical questions corpus
consists of 851
posts per class (1,702 total posts). Table \ref{table:rqs} shows some examples of rhetorical
questions and self-answering from our corpus. 

\vspace{.05in}
\noindent{\bf Hyperbole.} Hyperbole (Hyp) has been studied as an
independent form of figurative language, that can coincide with ironic
intent \cite{McCarthy2004,CanoMora2009}, and previous computational
work on sarcasm typically includes features to capture hyperbole
\cite{Reyesetal13}. \newcite{KreuzRoberts95} describe a standard frame
for hyperbole in English where an adverb modifies an extreme, positive
adjective, e.g. {\it "That was {\bf absolutely amazing}!"} or {\it "That
  was {\bf simply the most incredible} dining experience in my entire
  life."}

\newcite{ColstonObrien00b} provide a theoretical framework that
explains why hyperbole is so strongly associated with sarcasm.
Hyperbole exaggerates the literal situation, introducing a discrepancy
between the "truth" and what is said, as a matter of degree. A key
observation is that this is a type of contrast
\cite{ColstonKeller98,ColstonObrien00b}. In their framework:
\begin{itemize}
\item  An event or situation evokes a scale;
\vspace{-.15in}
\item  An event can be placed on that scale;
\vspace{-.15in}
\item  The utterance about the event {\bf contrasts} with actual scale placement.
\end{itemize}

\begin{figure}[h]
\begin{center}
    \includegraphics[width=2.2in]{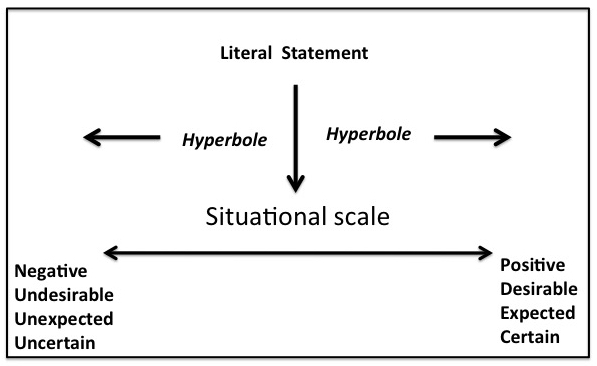}
  \caption{Hyperbole shifts the strength of what is said from literal to extreme negative
or positive (Colston and O'Brien, 2000)}
  \label{colston-fig}
  \end{center}
\end{figure}

Fig. \ref{colston-fig} illustrates that the scales that can be evoked
range from negative to positive, undesirable to desirable, unexpected
to expected and certain to uncertain.  Hyperbole moves the strength of
an assertion further up or down the scale from the literal meaning,
the degree of movement corresponds to the degree of
contrast.  Depending on what they modify, adverbial intensifiers like
{\it totally, absolutely, incredibly} shift the strength of the
assertion to extreme negative or positive. 

\begin{table}[h!t]
\begin{scriptsize} 
\centering
\ra{1.3}
\begin{tabular}{@{}p{7.5cm}@{}}\toprule
\bf \cellcolor[gray]{0.9} Hyperbole with Intensifiers \\ \midrule
Wow! {\bf I am soooooooo amazed} by your come back skills... 
another epic fail! \\ \hline
My goodness...{\bf i'm utterly amazed} at the number of 
men out there that are so willing to decide how a woman 
should use her own body! \\ 
\toprule
Oh do go on. {\bf I am so impressed} by your 'intellectuall' 
argument. pfft. \\ \hline
{\bf I am very impressed} with your ability to copy and paste 
links now what this proves about what you know about 
it is still unproven. \\ 
\bottomrule
\end{tabular}
\caption{Examples of Hyperbole and the Effects of Intensifiers \\ \label{hyperbole}}
\end{scriptsize}
\end{table}

Table~\ref{hyperbole} shows examples of hyperbole from our corpus,
showcasing the effect that intensifiers have in terms of strengthening
the emotional evaluation of the response.  To construct a balanced
corpus of sarcastic and not-sarcastic utterances with hyperbole, we
developed a number of patterns based on the literature and our
observations of the generic corpus. The patterns, number matches on
the whole corpus, the numbers we had annotated and the percent that
are sarcastic after annotation are summarized in
Table~\ref{table:cue-annotation}. Again, we extract a small subset of
examples from our Gen corpus (30 per class), and supplement them with posts
that contain our hyperbole cues (considering them sarcastic if they received at least 3/5 sarcastic labels, not-sarcastic otherwise). The
final hyperbole dataset consists of 582 posts per class (1,164 posts in total). 

To recap, Table \ref{new-corpus-stats}
summarizes the total number of posts for each subset of our final corpus.


\section{Learning Experiments}
\label{ml-sec}

Our primary goal is not to optimize classification results, but to
explore how results vary across different subcorpora and corpus
properties. We also aim to demonstrate that the quality of our corpus
makes it more straightforward to achieve high classification performance. We apply both supervised learning using SVM (from Scikit-Learn \cite{Pedregosaetal11}) and
weakly-supervised linguistic pattern learning using AutoSlog-TS
\cite{Riloff96}. These reveal different
aspects of the corpus. 
\vspace{.05in}

\noindent{\bf Supervised Learning.} We restrict our supervised
experiments to a default linear SVM learner with Stochastic Gradient
Descent (SGD) training and L2 regularization, available in the SciKit-Learn toolkit \cite{Pedregosaetal11}.  We use 10-fold
cross-validation, and only two types of features: n-grams and Word2Vec
word embeddings.  We expect Word2Vec to be able to capture semantic
generalizations that n-grams do not \cite{Socheretal13,Lietal16}.  The
n-gram features include unigrams, bigrams, and trigrams, including
sequences of punctuation (for example, ellipses or "!!!"), and emoticons.  We use
GoogleNews Word2Vec features ~\cite{Mikolovetal13}.\footnote{We test
our own custom 300-dimensional embeddings created for the
dialogic domain using the Gensim library~\cite{gensim}, and a very
large corpus of user-generated dialogue. While this custom model works
well for other tasks on IAC 2.0, it did not work well for sarcasm
classification, so we do not discuss it further.}

%
%

\begin{table}[tbh]
\begin{small}
\centering
\ra{1.3}
\begin{tabular}{@{}p{.6cm}|p{1.2cm}|P{0.7cm}|P{1cm}|P{1cm}|P{1cm}@{}}\toprule
\bf Form & \bf Features & \bf Class & \bf P & \bf R  & \bf F  \\ \midrule 
\bf Gen & N-Grams & $S$ & 0.73 & 0.70 & 0.72 \\
& & $NS$ & 0.71 & 0.75 & {\bf 0.73} \\ \midrule
& W2V & $S$ & 0.71 & 0.77 & {\bf 0.74} \\
& & $NS$ & 0.75 & 0.69 & 0.72 \\ 
\hline               \hline              
\bf RQ & N-Grams & $S$ & 0.71 & 0.68 & {\bf 0.70} \\
& & $NS$ & 0.70 & 0.73 & {\bf 0.71} \\ \midrule
& W2V & $S$ & 0.67 & 0.72 & 0.69 \\
& & $NS$ & 0.70 & 0.64 & 0.67 \\ 
\hline            \hline              
\bf Hyp & N-Grams & $S$ & 0.68 & 0.63 & {\bf 0.65} \\
& & $NS$ & 0.66 & 0.71 & {\bf 0.68} \\ \midrule
& W2V & $S$ & 0.57 & 0.56 & 0.57 \\
& & $NS$ & 0.57 & 0.59 & 0.58 \\ 
\hline               
 \end{tabular}
\end{small}
\vspace{-.1in}
\caption{\label{supervised-res} Supervised Learning Results for Generic (Gen: 3,260 posts per class), Rhetorical Questions (RQ: 851 posts per class) and Hyperbole (Hyp: 582 posts per class)\\}
\vspace{-.1in}
\end{table}

Table \ref{supervised-res} summarizes the results of our supervised
learning experiments on our datasets using 10-fold cross validation. The data is balanced evenly
between the {\sc sarcastic} and {\sc not-sarcastic} classes, and the best F-Measures for each
class are shown in bold.  The default W2V model, (trained on Google News),
gives the best overall F-measure of 0.74 on the Gen corpus for the
{\sc sarcastic} class, while n-grams give the best {\sc not-sarcastic}
F-measure of 0.73. Both of these results are higher F than previously reported
for classifying sarcasm in dialogue, and we might expect that feature
engineering could yield even greater performance.

\begin{figure}[h!t]
\begin{center}
    \includegraphics[width=3in]{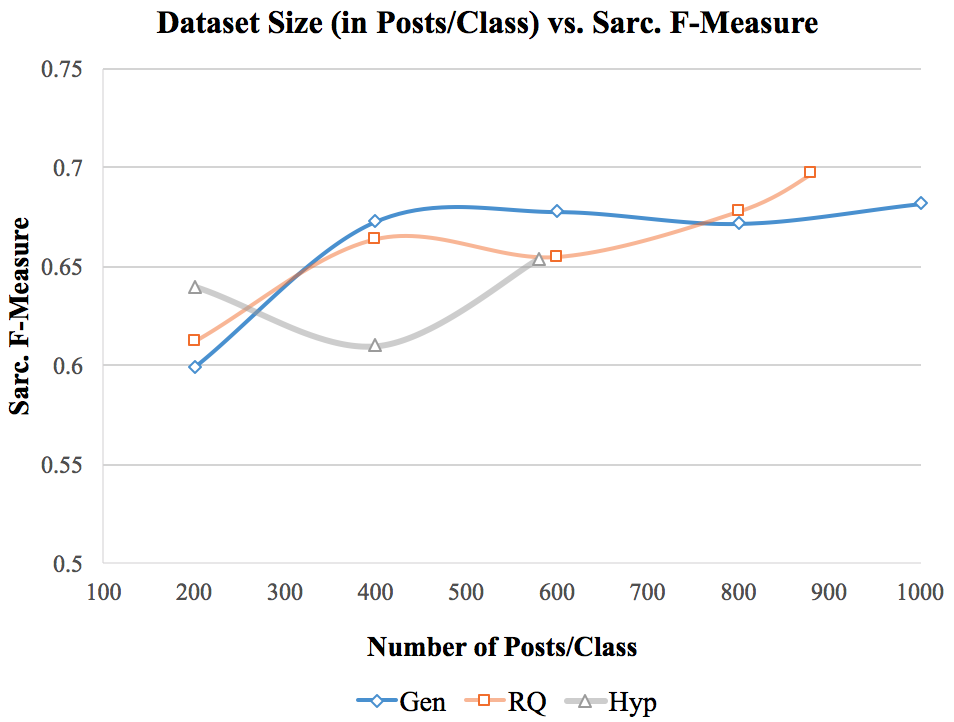}
    \caption{Plot of Dataset size (x-axis) vs Sarc. F-Measure (y-axis) for the three subcorpora, with n-gram features}
  \label{all-three}
  \end{center}
\vspace{-.2in}
\end{figure} 

On the RQ corpus, n-grams provide the best F-measure for {\sc
  sarcastic} at 0.70 and {\sc not-sarcastic} at 0.71. Although W2V
performs well, the n-gram model includes features involving repeated
punctuation and emoticons, which the W2V model excludes.  Punctuation 
and emoticons are often used as distinctive feature of sarcasm (i.e. {\it "Oh, really?!?!"}, {\it [emoticon-rolleyes]}).

For the Hyp corpus, the best F-measure for both the {\sc sarcastic}
and {\sc not-sarcastic} classes again comes from n-grams, with
F-measures of 0.65 and 0.68 respectively. It is interesting to note that
the overall results of the Hyp data are lower than those for Gen and RQs,
likely due to the smaller size of the Hyp dataset.


To examine the effect of dataset size, we compare F-measure (using the
same 10-fold cross-validation setup) for each dataset while holding the number
of posts per class constant. Figure \ref{all-three} shows the performance
of each of the Gen, RQ, and Hyp datasets at intervals of 100
posts per class (up to the maximum size of 582 posts per class for Hyp, and
851 posts per class for RQ). From the graph, we can see that as a general
trend, the datasets benefit from larger dataset sizes. Interestingly,
the results for the RQ dataset are very comparable to those of Gen. The Gen
dataset eventually gets the highest sarcastic F-measure (0.74) at its
full dataset size of 3,260 posts per class. 

\vspace{.05in}
\noindent{\bf Weakly-Supervised Learning.}  AutoSlog-TS is a weakly
supervised pattern learner that only requires training documents labeled
broadly as {\sc sarcastic} or {\sc not-sarcastic}. AutoSlog-TS uses a
set of syntactic templates to define different types of linguistic
expressions.  The left-hand side of Table~\ref{pattern-types} lists
each pattern template and the right-hand side illustrates a specific
lexico-syntactic pattern ({\bf in bold}) that represents an instantiation of each general pattern
template for learning sarcastic patterns in our
data.\footnote{The examples are shown as general expressions for
  readability, but the actual patterns must match the syntactic
  constraints associated with the pattern template.}  In addition to
these 17 templates, we added patterns to AutoSlog for adjective-noun,
adverb-adjective and adjective-adjective, because these patterns are
frequent in hyperbolic sarcastic utterances.

The examples in Table~\ref{pattern-types}
show that Colston's notion of contrast shows up in many
learned patterns, and that the source of the contrast is {\bf highly
variable}. For example, Row 1 implies a contrast with a set of people
who are not {\it your mother}. Row 5 contrasts what {\it you were asked}
with {\it what you've (just) done}. Row 10 contrasts {\it chapter 12}
and {\it chapter 13} \cite{Hirschberg85}. Row 11 contrasts {\it what I am allowed}
vs. what {\it you have to do}.

\begin{table}[t!bh]
\begin{scriptsize}
\centering
\ra{1.3}
\begin{tabular}{@{}p{0.05in}|p{.86in}|p{1.75in}@{}}\toprule
\bf \cellcolor[gray]{0.9} & \bf \cellcolor[gray]{0.9} Pattern Template & \bf \cellcolor[gray]{0.9} Example Instantiations \\ \midrule
1 & $<$subj$>$ PassVP & Go tell your mother, {\bf $<$she$>$ might be interested} in your fulminations. \\ \midrule
 2 &   $<$subj$>$ ActVP        & Oh my goodness. This is a trick called semantics. {\bf $<$I$>$ guess} you got sucked in.\\ \midrule
3&    $<$subj$>$ ActVP Dobj   & yet {\bf $<$I$>$ do nothing} to prevent the situation  \\ \midrule
4&    $<$subj$>$ ActInfVP & I guess {\bf $<$I$>$ need to check} what website I am in  \\ \midrule
5&    $<$subj$>$ PassInfVP & {\bf $<$You$>$ were asked to give} us your explanation of evolution. So far you've just ... \\ \midrule 
6&    $<$subj$>$ AuxVP Dobj & Fortunately {\bf $<$you$>$  have the ability} to ...
\\ \midrule
7&    $<$subj$>$ AuxVP Adj & Or do you think that {\bf $<$nothing$>$ is capable} of undermining the institution of marriage? \\ \midrule \midrule
8&    ActVP $<$dobj$>$ & Oh yes, I {\bf know $<$everything$>$} that [...] \\ \midrule
9&    InfVP $<$dobj$>$ & Good idea except we do not have {\bf to elect $<$him$>$} to any post... just send him over there.\\ \midrule
10&    ActInfVP $<$dobj$>$ & {\bf Try to read $<$chptr 13$>$} before chptr 12, it will help you out. \\ \midrule
 11 &   PassInfVP $<$dobj$>$ & i love it when people do this. 'you have to prove everything you say, but i {\bf am allowed to }simply {\bf make $<$assertions$>$} and it's your job to show i'm wrong.'\\ \midrule
 12&       Subj AuxVP $<$dobj$>$   & So your {\bf answer} [then] {\bf is $<$nothing$>$}... \\ 
    \midrule \midrule
13&    NP Prep $<$np$>$  &  There are {\bf MILLIONS of $<$people$>$} saying all sorts of stupid things about the president.\\ \midrule
14&    ActVP Prep $<$np$>$ & My pyramidal tinfoil hat is an antenna for knowledge and truth. It reflects  idiocy and dumbness into deep space. You still have not {\bf admitted to $<$your error$>$} \\ \midrule
15&    PassVP Prep $<$np$>$ & Likelihood is that they will have to  {\bf be left alone for $<$a few months$>$} [...] Sigh, I wonder if ignorance really is blissful. \\  \midrule
16&    InfVP Prep $<$np$>$ & I masquerade as an atheist and a 6-day creationist at the same time to try {\bf to appeal to $<$a wider audience$>$}. \\ \midrule
17&    $<$possessive$>$ NP & O.K. let's play {\bf $<$your$>$ game}. \\ 
 \bottomrule
 \end{tabular}
\end{scriptsize}
\vspace{-.1in}
  \caption{\label{pattern-types} AutoSlog-TS Templates and Example Instantiations}
\vspace{-.1in}
\end{table}

AutoSlog-TS computes statistics on
the strength of association of each pattern with each class,
i.e. P({\sc sarcastic} $\mid$ $p$) and P({\sc not-sarcastic} $\mid$
$p$), along with the pattern's overall frequency. We define two tuning parameters for each
class: $\theta_f$, the frequency with which a pattern occurs,
$\theta_p$, the probability with which a pattern is associated with
the given class. We do a grid-search, testing the performance of our
patterns thresholds from $\theta_f$ = \{2-6\} in intervals of
1, $\theta_p$=\{0.60-0.85\} in intervals of 0.05. Once we extract the subset
of patterns passing our thresholds, we search for these patterns in
the posts in our development set, classifying a post as a given class
if it contains $\theta_n$=\{1, 2, 3\} of the thresholded patterns.
For more detail, see \cite{Riloff96,Orabyetal15}.

\begin{table*}[t!h]
\begin{scriptsize}
\ra{1.3}
\begin{tabular}{@{}P{1cm}|P{1cm}|P{5cm}|P{7.5cm}@{}}
\toprule
\bf Prob. & \bf Freq. & {\bf Pattern and Text Match} & \textbf{Sample Post}\\ \hline
\multicolumn{4}{c}{\bf \cellcolor[gray]{0.9}  Sarcastic Example Patterns}  \\\hline
\bf 1.00 & 8 & Adv Adv (\texttt{AH YES}) &  \textbf{Ah yes}, your diversionary tactics. \\ 
\bf 0.91 & 11 & Adv Adv ({\tt THEN AGAIN}) & But {\bf then again}, you become what you hate [...] \\
\bf 0.83 & 36 & ActVP Prep $<$NP$>$ (\texttt{THANKS FOR}) &  \textbf{Thanks for} missing the point. \\

\bf 0.80 & 20 & ActVP $<$dobj$>$ ({\tt TEACH}) & {\bf Teach} the science in class and if that presents a problem [...] \\
\bf 0.80 & 10 & InfVP $<$dobj$>$ ({\tt ANSWER}) & I think you need to {\bf answer} the same question [...] \\
\bf 0.79 & 114 & $<$subj$>$ActVp ({\tt GUESS}) & So then I {\bf guess} you could also debate that algebra serves no purpose \\
\bf 0.78 & 18 & ActVP $<$dobj$>$ ({\tt IGNORE}) & Excellent {\bf ignore} the issue at hand and give no suggestion \\
\bf 0.74 & 27 & Adv Adv (\texttt{ONCE AGAIN}) & you attempt to \textbf{once again} change the subject \\
\bf 0.71 & 35 & Adj Noun (\texttt{GOOD IDEA}) & ...especially since you think everything is a \textbf{good idea} \\

  \hline
\multicolumn{4}{c}{\bf \cellcolor[gray]{0.9}  Not-Sarcastic Example Patterns}  \\ \hline
\bf 0.92 & 25 & Adj Noun (\texttt{SECOND AMENDMENT}) & the nature of the \textbf{Second Amendment} \\
\bf 0.90 & 10 & Np Prep $<$NP$>$ (\texttt{PROBABILITY OF}) & the \textbf{probability of} [...] in some organism\\
\bf 0.88 & 42 & ActVP $<$dobj$>$ (\texttt{SUPPORT}) & I really do not {\bf support} rule by the very, very few \\ 
\bf 0.84 & 32 & Np Prep $<$NP$>$ (\texttt{EVIDENCE FOR}) & We have no more \textbf{evidence for} one than the other. \\
\bf 0.79 & 44  & Np Prep (\texttt{THEORY OF}) &  [...] supports the \textbf{theory of} evolution [...]\\
\bf 0.78 & 64 & Np Prep $<$NP$>$ (\texttt{NUMBER OF}) & minor differences in a limited {\bf number of} primative organisms\\
\bf 0.76 & 46 & Adj Noun (\texttt{NO EVIDENCE}) & And there is {\bf no evidence }of anything other than material processes\\

\bf 0.75 & 41 & Np Prep $<$NP$>$ (\texttt{MAJORITY OF}) & The {\bf majority of} criminals don't want to deal with trouble. \\
\bf 0.72 & 25 & ActVP $<$dobj$>$ (\texttt{EXPLAIN}) &  [...] it does not \textbf{explain} the away the whole shift in the numbers [..]\\
 \bottomrule
 
 \end{tabular}
\end{scriptsize}
\caption{\label{table:high-patterns} Examples of Characteristic Patterns for Gen using AutoSlog-TS Templates\\}
\vspace{-.1in}
\end{table*}

\begin{figure}[h!]
\begin{center}
    \includegraphics[width=3.1in]{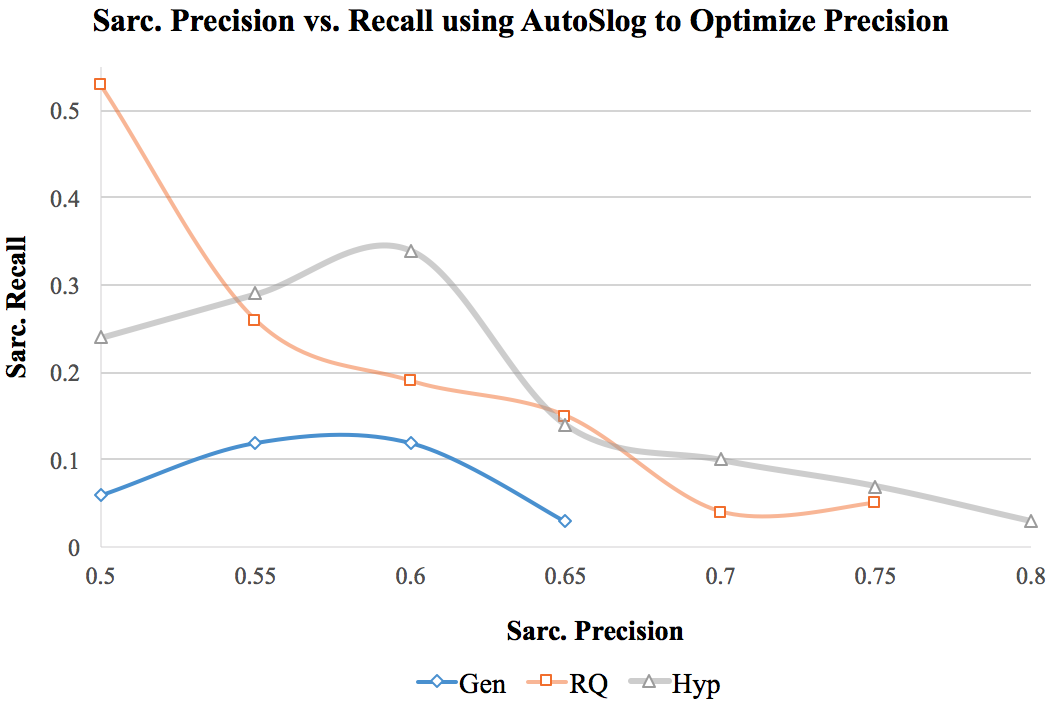}
    \caption{Plot of Precision (x-axis) vs Recall (y-axis) for three subcorpora with AutoSlog-TS parameters, aimed at optimizing precision}
  \label{all-three-P}
  \end{center}
\vspace{-.2in}
\end{figure}

An advantage of AutoSlog-TS is that it supports systematic exploration
of recall and precision tradeoffs, by selecting pattern sets using
different parameters. The parameters have to be tuned on a training
set, so we divide each dataset into 80\% training and 20\% test.
Figure~\ref{all-three-P} shows the precision (x-axis) vs. recall
(y-axis) tradeoffs on the test set, when optimizing our three parameters for
precision. Interestingly, the subcorpora for RQ and Hyp can get higher
precision than is possible for Gen. When precision is fixed at
0.75, the recall for RQ is 0.07 and the recall for Hyp is 0.08. This
recall is low, but given that each retrieved post provides
multiple cues, and that datasets on the web are
huge, these P values make it possible to bootstrap these two
classes in future.

\section{Linguistic Analysis}
\label{ling-anal}

Here we aim to provide a linguistic characterization
of the differences between the sarcastic and
the not-sarcastic classes. 
We use the AutoSlog-TS pattern learner to generate patterns
automatically, and the Stanford dependency parser to examine
relationships between arguments \cite{Riloff96,Manningetal14}.  Table
\ref{new-patterns} shows the number of sarcastic patterns we extract
with AutoSlog-TS, with a frequency of at least 2 and a probability of
at least 0.75 for each corpus.  We learn many
novel lexico-syntactic cue patterns that are not the regex that we
search for. We discuss specific novel learned
patterns for each class below.\\

\noindent{\bf Generic Sarcasm.} We first examine
the different patterns learned on the Gen dataset.  Table
\ref{table:high-patterns} show examples of extracted patterns for each
class. We observe that the {\sc not-sarcastic} patterns appear to
capture technical and scientific language, while the {\sc sarcastic}
patterns tend to capture subjective language that is not
topic-specific. We observe an abundance of adjective and adverb patterns
for the sarcastic class, although we do not use adjective and
adverb patterns in our regex retrieval method. Instead, such cues
co-occur with the cues we search for, expanding our pattern inventory
as we show in Table \ref{new-patterns}. 

\begin{table}[h!]
\begin{scriptsize}
\centering
\ra{1.3}
\begin{tabular}{@{}P{3.8cm}|P{1.5cm}|P{1.5cm}@{}}\toprule
\bf Dataset & \bf  \# Sarc Patterns & \bf \# NotSarc Patterns \\ \midrule 
Generic (Gen)  & 1,316 & 3,556 \\ \midrule 
Rhetorical Questions (RQ)  & 671 & 1,000 \\ \midrule 
Hyperbole (Hyp) & 411 & 527 \\ 
 \bottomrule
 \end{tabular}
\end{scriptsize}
\vspace{-.1in}
\caption{\label{new-patterns} Total number of patterns passing threshold of Freq $ \ge 2$, Prob $ \ge 0.75$\\}
\end{table}

\begin{table}[h]
\begin{scriptsize}
\centering
\ra{1.3}
\begin{tabular}{@{}P{2.8cm}|P{4.4cm}@{}}\toprule
\bf Relation & \bf  Rhetorical Question \\ \midrule 
{\tt realize(you, human)}  & Uhm, {\bf you do realize that humans} and chimps are not the same things as dogs, cats, horses, and sharks ... right? \\ \midrule 
{\tt recognize(you)}  & Do {\bf you recognize} that babies grow and live inside women? \\ \midrule 
{\tt not read(you)}  & Are you blind, or {\bf can't you read}? \\ \midrule 
{\tt get(information)}  & Have you ever considered {\bf getting scientific information} from a scientific source? \\ \midrule 
{\tt have(education)} & And you claim to have an education? \\ \midrule
{\tt not have(dummy, problem)} & If these {\bf dummies don't have a problem} with information increasing, but do have a problem with beneficial information increasing, don't you think there is a problem? \\
 \bottomrule
 \end{tabular}
\end{scriptsize}
\vspace{-.1in}
\caption{\label{rq-relations} Attacks on Mental Ability in RQs}
\end{table}

\noindent{\bf Rhetorical Questions.}
We notice that while the {\sc not-sarcastic} patterns
generated for RQs are similar to the topic-specific {\sc not-sarcastic} patterns we find in
the general dataset, there are some interesting features of the {\sc sarcastic} patterns that 
are more unique to the RQs.

Many of our sarcastic questions focus specifically on attacks on
the mental abilities of the addressee. This
generalization is made clear when we extract and analyze the verb, subject,
and object arguments using the Stanford dependency parser
\cite{Manningetal14} for the {\it questions} in the RQ dataset. Table
\ref{rq-relations} shows a few examples of the relations we extract.

\vspace{.05in}
\noindent{\bf Hyperbole.}  One common pattern for hyperbole involves
adverbs and adjectives, as noted above. We did not use this pattern to
retrieve hyperbole, but because each hyperbolic sarcastic utterance
contains multiple cues, we learn an expanded class of patterns for
hyperbole. Table~\ref{new-hyp-adv-tab} illustrates some of the new
adverb adjective patterns that are frequent, high-precision indicators
of sarcasm. 

We learn a number of verbal patterns that we
had not previously associated with hyperbole, as shown in Table~\ref{new-hyp-verb-tab}.  Interestingly, many of these instantiate the observations of \newcite{CanoMora2009} on hyperbole and its
related semantic fields: creating contrast by
exclusion, e.g. {\it no limit} and {\it no way}, or by expanding a
predicated class, e.g. {\it everyone knows}. Many of them are also
contrastive. Table~\ref{new-hyp-adv-tab} shows just a few examples, such as {\it though it in no way} and {\it so much knowledge}. 

\begin{table}[t!bh]
\begin{scriptsize}
\begin{center}
\begin{tabular}{@{}P{2.0cm}|P{.4cm}|P{4cm}@{}}\toprule
\bf Pattern & \bf Freq  & \bf Example \\ \hline 
    \texttt{no way} & 4 & that is a pretty impresive education you are working on (though it in {\bf no way} makes you a shoe in for any political position). \\ \hline
   \texttt{so much} & 17 & but nooooooo we are launching missiles on libia thats solves alot ....  because we gained {\bf so much} knowledge and learned from our mistakes \\ \hline
   \texttt{oh dear}  & 12 & {\bf oh dear}, he already added to the gene pool\\ \hline
   \texttt{how much }  & 8 & you have no idea {\bf how much} of a hippocrit you are, do you\\ \hline
 \texttt{exactly what} & 5  & simone, {\bf exactly what} is a gun-loving fool anyway, other than something you...\\ \hline
\end{tabular}
\end{center}
\end{scriptsize}
\vspace{-.1in}
\caption{Adverb Adjective Cues in Hyperbole \label{new-hyp-adv-tab}}
\end{table}

\begin{table}[h!]
\begin{scriptsize}
\begin{center}
\begin{tabular}{P{2.0cm}|P{.4cm}|P{4cm}}\toprule
\bf Pattern & \bf Freq & \bf Example \\ \hline 
     \texttt{i bet} &  9  & {\bf i bet} there is a university thesis in there somewhere \\ \hline
     \texttt{you don't see} & 7 & {\bf you don't see} us driving in a horse and carriage, do you  \\ \hline
    \texttt{everyone knows} & 9 & {\bf everyone knows} blacks commit more crime than other races  \\ \hline
 \texttt{I wonder}  & 5  & hmm {\bf i wonder} ware the hot bed for violent christian extremists is \\ \hline
   \texttt{you trying} & 7 & if {\bf you are seriously trying} to prove your god by comparing real life things with fictional things, then yes,  you have proved your god is fictional \\ 
 \bottomrule
 \end{tabular}
\end{center}
\end{scriptsize}
\vspace{-.1in}
\caption{Verb Patterns in Hyperbole \label{new-hyp-verb-tab}}
\vspace{-.1in}
\end{table}

\section{Conclusion and Future Work}
\label{conc-sec}

We have developed a large scale, highly diverse
corpus of sarcasm using a combination of linguistic analysis
and crowd-sourced annotation. We use filtering methods to
skew the distribution of sarcasm in posts to be annotated
to 20-31\%, much higher than the estimated 12\% distribution
of sarcasm in online debate forums. We note that when
using Mechanical Turk for sarcasm annotation, it is possible that
the level of agreement signals how lexically-signaled the sarcasm is, so
we settle on a conservative threshold (at least 6 out of 9 annotators agreeing that a post is sarcastic) to ensure the quality of our annotations. 

We operationalize lexico-syntactic
cues prevalent in sarcasm, finding cues that are highly indicative of sarcasm, with ratios up to 87\%. Our final corpus consists of data representing generic sarcasm, rhetorical questions,
and hyperbole.

We conduct supervised learning experiments to highlight the quality of
our corpus, achieving a best F of 0.74 using very simple feature sets. We use
weakly-supervised learning to show that we can also achieve high
precision (albeit with a low recall) for our rhetorical questions and
hyperbole datasets; much higher than the best precision that is
possible for the Generic dataset. These high precision values
may be used for bootstrapping these two classes in
the future.

We also present qualitative analysis of the different characteristics of rhetorical questions and hyperbole in sarcastic acts,
and of the distinctions between sarcastic/not-sarcastic cues in
generic sarcasm data. Our analysis shows that the forms of sarcasm and
its underlying semantic contrast in dialogue are highly
diverse. 

In future work, we will focus on feature engineering to
improve results on the task of sarcasm classification for both our generic data
and subclasses. We will also begin to explore evaluation on real-world data distributions, where
the ratio of sarcastic/not-sarcastic posts is inherently unbalanced. 
As we continue our analysis of the generic and fine-grained
categories of sarcasm, we aim to better characterize and model the great diversity of sarcasm in dialogue.
\section*{Acknowledgments}
\label{ack}
This work was funded by NSF CISE RI 1302668,
under the Robust Intelligence Program. 

\bibliography{../../nl}
\bibliographystyle{acl2016}

\end{document}